\documentclass[letterpaper, 10 pt, conference]{ieeeconf}  

\IEEEoverridecommandlockouts                              

\usepackage{tcolorbox}
\usepackage{lipsum}
\usepackage{graphicx}
\usepackage{makecell}
\usepackage{cite}
\usepackage{amsmath,amssymb,amsfonts}
\usepackage{algorithm}
\usepackage{algorithmic}
\usepackage{comment}
\usepackage{caption}
\usepackage{subcaption}
\usepackage{multirow}
\usepackage{float}    
\usepackage{placeins} 
\usepackage{hyperref}
\usepackage{booktabs}
\usepackage{svg}
\usepackage[acronym]{glossaries}

\newacronym{adas}{ADAS}{Advanced Driver-Assistance Systems}
\newacronym{ads}{ADS}{Automated Driving Systems}
\newacronym{acc}{ACC}{Adaptive Cruise Control}
\newacronym{aeb}{AEB}{Autonomous Emergency Braking}
\newacronym{ctrv}{CTRV}{Constant Turn Rate and Velocity}

\newacronym{llm}{LLM}{Large Language Model}

\newacronym{koi}{KeyIdent}{Key Object Identification}

\newacronym{id}{ID}{In-Distribution}
\newacronym{ood}{OOD}{Out-of-Distribution}

\newacronym{tp}{TP}{True Positive}
\newacronym{tpr}{TPR}{True Positive Rate}
\newacronym{fp}{FP}{False Positive}
\newacronym{fn}{FN}{False Negative}

\newacronym{ttc}{TTC}{Time to Collision}

\newacronym{rag}{RAG}{Retrieval-Augmented Generation}

\newacronym{uuid}{UUID}{Unique Identifier}

\title{\LARGE \bf Why Braking? Scenario Extraction and Reasoning Utilizing LLM}
\author{Yin Wu$^{1, 2}$, Daniel Slieter$^{1}$, Vivek Subramanian$^{1}$,\\ Ahmed Abouelazm$^{3}$, Robin Bohn$^{1}$, and J. Marius Zöllner$^{2, 3}$
\thanks{$^{1}$Authors are with the CARIAD SE, Germany \newline {\tt\small firstname.lastname@cariad.technology}}
\thanks{$^{2}$Authors are with the Karlsruhe Institute of Technology, Germany}%
\thanks{$^{3}$Authors are with the FZI Research Center for Information Technology, Germany}%
}

\begin{document}
\maketitle
\thispagestyle{empty}
\pagestyle{empty}

\begin{abstract}
    The growing number of ADAS-equipped vehicles has led to a dramatic increase in driving data, yet most of them capture routine driving behavior. Identifying and understanding safety-critical corner cases within this vast dataset remains a significant challenge. Braking events are particularly indicative of potentially hazardous situations, motivating the central question of our research: \textit{Why does a vehicle brake?} Existing approaches primarily rely on rule-based heuristics to retrieve target scenarios using predefined condition filters. While effective in simple environments such as highways, these methods lack generalization in complex urban settings. In this paper, we propose a novel framework that leverages \gls{llm} for scenario understanding and reasoning. Our method bridges the gap between low-level numerical signals and natural language descriptions, enabling \gls{llm} to interpret and classify driving scenarios. We propose a dual-path scenario retrieval that supports both category-based search for known scenarios and embedding-based retrieval for unknown \gls{ood} scenarios. To facilitate evaluation, we curate scenario annotations on the Argoverse 2 Sensor Dataset. Experimental results show that our method outperforms rule-based baselines and generalizes well to \gls{ood} scenarios.
    
\end{abstract}

\section{Introduction}
\label{sec:Introduction}

\begin{figure}[ht]
    \centering
    \includegraphics[width=\linewidth]{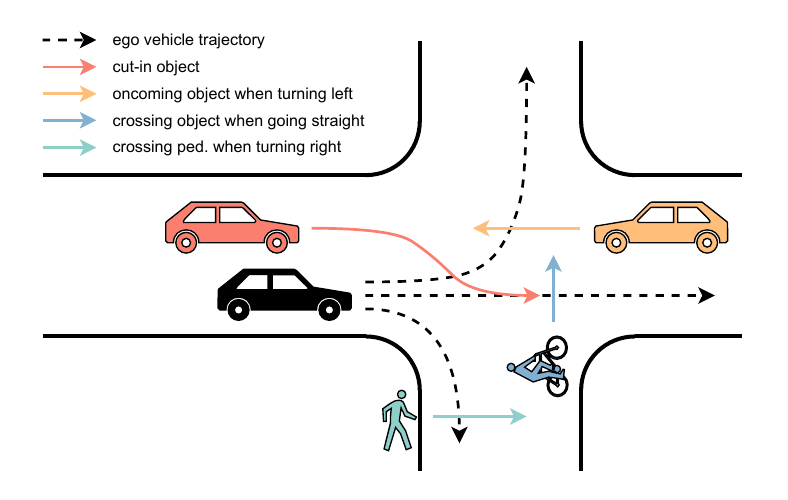}
    \caption{Four representative scenarios where the black ego vehicle brakes to avoid obstacles: waiting for a yellow oncoming vehicle when turning left, yielding to a green pedestrian when turning right, slowing down for a red cut-in vehicle or a crossing blue bicycle when going straight.}
    \label{fig:sc_cases}
\end{figure}

With the growing deployment of \gls{adas}-equipped vehicles, the volume of driving data collected has increased dramatically. According to~\cite{krzanich2016data}, an autonomous vehicle produces about 4000 GB of data a day, depending on its sensor suite, such as cameras, LiDAR, radar, and BUS-Systems. While most of the data captures routine driving behaviors, only a small subset of data is relevant for functional validation. Among various driving behaviors, braking is a particularly salient signal that often correlates with potentially hazardous situations, such as obstacle avoidance, cut-ins, or yielding at intersections. Understanding why a vehicle brakes can thus reveal meaningful corner cases and enhance the coverage of scenario-based assessments. Fig.~\ref{fig:sc_cases} illustrates four representative examples in which the ego vehicle brakes: a yellow oncoming vehicle during a left turn, a green pedestrian crossing during a right turn, a red vehicle cutting in, and a bicycle crossing while the ego proceeds straight. The ego vehicle's ADAS system must promptly brake in these four scenarios to avoid conflicts. 

While a braking signal is easy to detect, identifying its cause is not. This motivates the central question of this work: \textit{Why braking?} Our objective is to extract and reason about scenarios that causally lead to ego vehicle braking in a structured and explainable manner.


\begin{figure*}[htbp]
    \centering
    \includegraphics[width=\linewidth]{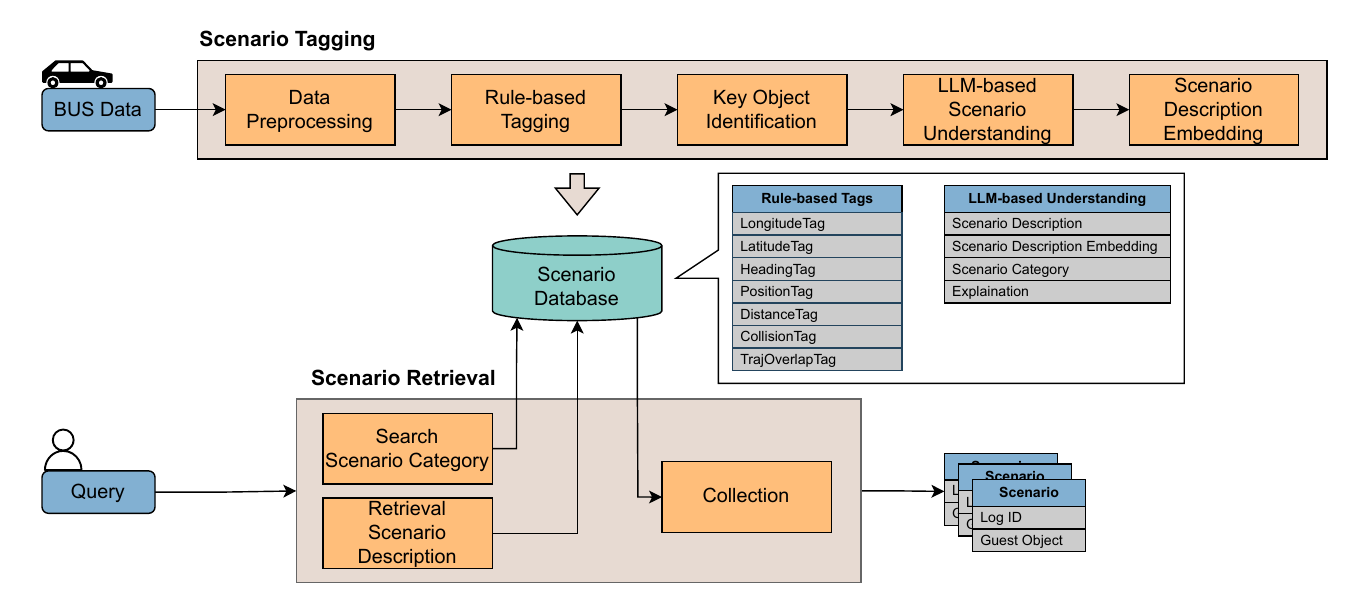}
    \caption{The scenario tagging and retrieval framework consists of several stages. The preprocessed raw driving data is first processed using rule-based methods to generate activity tags. The module \gls{koi} then detects candidate objects that potentially trigger the ego vehicle's braking behavior. For each ego–candidate object pair, the corresponding activity log is interpreted by an \gls{llm} to produce rephrased scenario descriptions, classify the scenario category, and give explanations. Subsequently, an embedding model generates a scenario description embedding. The enrichment information is stored in a scenario database, which is queried through a dual-path retrieval module.}
    \label{fig:model_structure}
\end{figure*}

To address this challenge, various scenario extraction approaches have been explored~\cite{de2020real, guo2023scenario, zhao2024chat2scenario, davidson2025refav}. A common paradigm is to use rule-based methods to tag agent activities~\cite{de2020real, guo2023scenario} and retrieve specific scenarios with carefully defined condition filters. However, defining these filter conditions is non-trivial: relaxing a filter may recover some false negatives (\gls{fn}) but can also introduce new false positives (\gls{fp}), and vice versa. Zhao et al.~\cite{zhao2024chat2scenario} propose using \gls{llm} to automatically generate filter conditions, yet this approach still faces similar trade-offs between recall and precision.


In this work, we move beyond using \gls{llm} solely as condition translators and instead leverage them for holistic scenario understanding. To bridge the gap between low-level numerical BUS data and high-level semantic reasoning, we first extract symbolic tags using the scenario tagging method proposed by Guo et al.~\cite{guo2023scenario}, and convert them into structured natural language descriptions. To improve efficiency and avoid irrelevant context, we first identify the key objects that are causally related to the ego vehicle’s braking event. For each identified object, we construct a scenario description that captures the interaction between the ego vehicle and the candidate. These descriptions are then fed into the \gls{llm} to perform rephrasing, contextual reasoning, and scenario classification. Subsequently, a text embedding model is used to obtain an embedding vector from the rephrased scenario descriptions. For the scenario retrieval stage, we propose a dual-path retrieval framework that supports retrieving scenarios either directly based on the \gls{llm}-generated scenario categories or via similarity search using embeddings derived from a scenario description query.

To evaluate our proposed method, we annotate 700 data logs on the training set of the Argoverse 2 Sensor Dataset~\cite{wilson2023argoverse} with scenario labels. Experimental results show that our LLM-based approach not only outperforms the rule-based baseline~\cite{guo2023scenario} but also demonstrates zero-shot generalization on \gls{ood} unseen scenarios.





\section{Related Work}

In this section, we review three research areas relevant to our proposed method: scenario-based validation for \gls{adas}, scenario extraction methodologies, and \gls{llm}-based approaches for driving scenario understanding.

\subsection{From Mileage-Based to Scenario-Based Validation}
The validation of \gls{adas} and \gls{ads} requires more than just accumulating driving distance. According to ~\cite{kalra2016driving}, demonstrating the safety of an ADS would require billions of miles of on-road testing, which is practically infeasible. Consequently, the industry has shifted towards scenario-based validation methodologies~\cite{scholtes20216}. In this paradigm, an \gls{ads} is tested against a curated catalog of critical and corner-case scenarios.

A key challenge lies in identifying these crucial scenarios. While synthetic scenario generation is a popular approach~\cite{nalic2020scenario,schutt20231001,rempe2022generating}, it may not fully capture the complexity and distribution of real-world events, potentially overlooking rare but critical corner cases. To address this, our work focuses on extracting meaningful scenarios directly from large-scale real-world driving data. Specifically, we target scenarios that cause the ego-vehicle to perform braking maneuvers, as these are directly relevant to \gls{adas} functions like \gls{acc}, \gls{aeb}, etc.

\subsection{Scenario Extraction from Real-World Data}
Efficiently mining scenarios from vast datasets is a significant research area. An early and effective paradigm is "tagging-and-searching," where raw data is pre-processed into a database with descriptive tags, enabling fast queries~\cite{de2020real}. Subsequent works have focused on building the interaction tag between agents and environments to allow for more precise scenario definitions~\cite{guo2023scenario}. More recently, \gls{llm} have been used to translate natural language queries into formal search filters. For instance, Chat2Scenario~\cite{zhao2024chat2scenario} and RefAV~\cite{davidson2025refav} leverage \gls{llm} to find scenarios matching complex textual descriptions. However, these approaches primarily use \gls{llm} as language translators, rather than fully exploiting their capacity for contextual understanding and reasoning.

\begin{table}[htbp]
\centering
\caption{Definition of activity tags and values.}
\label{tab:scenario_tags}
\resizebox{0.98\columnwidth}{!}{
\begin{tabular}{lll}
\toprule
\textbf{Tag Category} & \textbf{Category Description} & \textbf{Tag Name} \\
\midrule
\multirow{5}{*}{LongitudeTag} & \multirow{5}{*}{\parbox{3cm}{Describes the vehicle's longitudinal motion along its path.}} & CRUISING \\
 & & STANDING\_STILL \\
 & & ACCELERATING \\
 & & DECELERATING \\
 & & REVERSING \\
\midrule
\multirow{5}{*}{LatitudeTag} & \multirow{5}{*}{\parbox{3cm}{Describes the vehicle's orientation, including turns and lateral shifts relative to its path.}} & FACING\_FORWARD \\
 & & VEERING\_LEFT \\
 & & VEERING\_RIGHT \\
 & & TURNING\_LEFT \\
 & & TURNING\_RIGHT \\
\midrule
\multirow{3}{*}{CollisionTag} & \multirow{3}{*}{\parbox{3cm}{Describes the collision risk between two agents.}} & NO \\
 & & LOW \\
 & & HIGH \\
\midrule
\multirow{4}{*}{HeadingTag} & \multirow{4}{*}{\parbox{3cm}{Describes the relative heading angle of guest to ego vehicle.}} & SAME \\
 & & LEFT \\
 & & RIGHT \\
 & & OPPOSITE \\
\midrule
\multirow{8}{*}{PositionTag} & \multirow{8}{*}{\parbox{3cm}{Describes the relative position of guest to ego vehicle.}} & FRONT \\
 & & FRONT\_LEFT \\
 & & LEFT \\
 & & BACK\_LEFT \\
 & & BACK \\
 & & BACK\_RIGHT \\
 & & RIGHT \\
 & & FRONT\_RIGHT \\
\midrule
\multirow{4}{*}{DistanceTag} & \multirow{4}{*}{\parbox{3cm}{Describes the bounding box distance between two objects.}} & VERY\_CLOSE \\
 & & CLOSE \\
 & & MEDIUM \\
 & & FAR \\
 \midrule
\multirow{4}{*}{TrajOverlapTag} & \multirow{3}{*}{\parbox{3.3cm}{Describe the overlap risk of the aggregated bounding boxes of two objects over a future time.}} & NO \\
 & & LOW \\
 & & HIGH \\ \\
\bottomrule
\end{tabular}}
\end{table}

\subsection{Large Language Models for Semantic Scenario Understanding}

Modern \gls{llm} have demonstrated impressive reasoning and zero-shot capabilities across both textual and visual modalities. Recent works such as DriveLM~\cite{sima2024drivelm} and DriveVLM~\cite{tian2024drivevlm} integrate \gls{llm} into end-to-end autonomous driving systems, focusing primarily on single-scene understanding based on visual inputs. However, these models are not designed to directly interpret time-series driving data. For instance, a 10-second driving sequence recorded at 10 Hz contains 100 frames. Feeding all 100-frame scene descriptions into the \gls{llm} results in context overload, reduced relevance, and longer inference time.

To integrate \gls{llm} in downstream tasks, the prompting engineer plays an important role in enhancing the output quality without modifying their underlying parameters. Techniques such as chain-of-thought prompting~\cite{wei2022chain} has been shown to improve reasoning capabilities and reduce output variance. Liu et al.~\cite{liu2023pre} highlighted that explicitly specifying the role of the \gls{llm}, alongside enforcing structured output formats such as JSON, can significantly enhance output quality and consistency. We referred to these methods when designing our prompts.

\section{Methodology}

As shown in Fig.~\ref{fig:model_structure}, our proposed scenario tagging and retrieval framework consists of two main components: a two-stage scenario tagging module and a dual-path scenario retrieval module. The scenario tagging module processes and enriches the recorded driving BUS data. The data are first preprocessed using interpolation and smoothing algorithms. Then, a rule-based tagging module is applied to extract activity tags for each agent as well as interaction tags between the ego vehicle and surrounding agents. Subsequently, the \gls{koi} module filters out non-relevant objects to reduce redundancy in the following steps. Based on these tags, the numerical data are translated into structured natural language descriptions and fed into a \gls{llm} to generate high-level scenario understanding. The \gls{llm} output includes a rephrased scenario description, a scenario category (based on given scenario categories in the prompt), and an explanation for the classification. An embedding model will then calculate the embedding vector for the rephrased scenario description. Both the rule-based tags and the LLM-enhanced understanding are stored in a scenario database. During the scenario retrieval process, we employ a dual-path retrieval strategy, supporting both attribute-based search and similarity-based search. In the following subsections, we provide detailed descriptions of each submodule. We assume that the input BUS data has been preprocessed and includes stable track \gls{uuid}.


\subsection{Data Preprocessing}

The input of each data log includes the ego vehicle's odometry information ($x$, $y$ coordinates and yaw angle), and the perception output of detected objects, which includes their $x$, $y$, yaw angle, width, height, object category, and a tracking \gls{uuid}.

Although the input data are assumed to be refined before entering the system, they may still contain noise or missing entries. To mitigate these issues, we first apply linear interpolation within the valid lifespan of each object to fill in missing values. Then, a Savitzky-Golay smoothing algorithm~\cite{schafer2011savitzky} is employed to denoise the data while preserving sharp motion transitions. We specifically choose this method because other commonly used smoothing techniques, such as spline fitting, tend to over-smooth the data, suppressing short but critical events such as sudden braking, which may lead to the omission of important scenario patterns.

\subsection{Rule-based Scenario Tagging}
\label{sec:trajoverlap}

Following the work of~\cite{guo2023scenario}, each scenario log is annotated with two types of tags: actor activities and actor–interaction activities. A key distinction from prior work is that our system does not rely on HD maps. This design choice is made to enhance the system's applicability in real-world scenarios where HD maps are often unavailable or unreliable. A detailed definition of each activity tag and the associated values is provided in Table~\ref{tab:scenario_tags}.

For the specific tagging rules related to "LongitudeTag", "LatitudeTag", "CollisionTag", "HeadingTag", "PositionTag", and "DistanceTag", we refer readers to the original work~\cite{guo2023scenario}, where these rules are thoroughly described. To avoid redundancy, we do not repeat those details here.

\begin{figure}[htbp]
    \centering
    \begin{subfigure}[t]{0.22\textwidth}
        \centering
        \tcbset{colframe=black,boxrule=0.5pt,arc=0mm,boxsep=0mm,left=0mm,right=0mm,top=0mm,bottom=0mm}
        \begin{tcolorbox}
            \includegraphics[width=\textwidth]{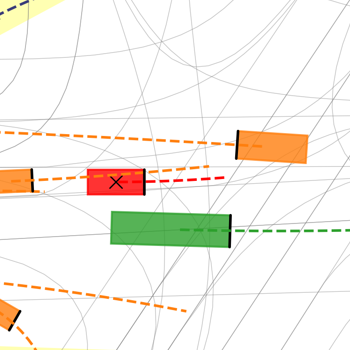}
        \end{tcolorbox}
        \caption{Predicted trajectory in 3 seconds based on \gls{ctrv} model. No collision risk detected for \textit{CollisionTag}}
        \label{fig:traj_exp_pred}
    \end{subfigure}
    \hfill
    \begin{subfigure}[t]{0.22\textwidth}
        \centering
        \tcbset{colframe=black,boxrule=0.5pt,arc=0mm,boxsep=0mm,left=0mm,right=0mm,top=0mm,bottom=0mm}
        \begin{tcolorbox}
            \includegraphics[width=\textwidth]{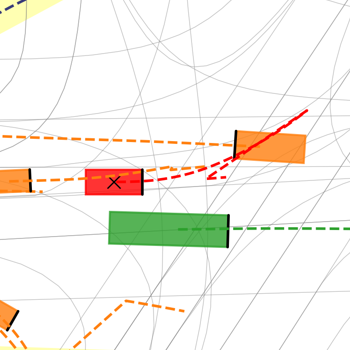}
        \end{tcolorbox}
        \caption{Future actual trajectory in 3 seconds. High overlap risk detected for \textit{TrajOverlapTag}}
        \label{fig:fc_approach_stop}
    \end{subfigure}

    \caption{Comparison between \textit{CollisionTag} (left) with \textit{TrajOverlapTag} (right). Red: ego vehicle; orange: guest vehicle; dash line: predicted / actual future trajectory.}
    \label{fig:exp_overlap}
\end{figure}

In addition to the existing tags, we introduce a new actor–interaction tag called \textit{TrajOverlapTag}. Unlike the \textit{CollisionTag} in~\cite{guo2023scenario}, which relies on motion prediction using a \gls{ctrv} model and \gls{ttc} estimation, our method directly leverages the ground-truth future trajectories recorded in the logs. For both the ego vehicle and a surrounding object, we aggregate all 2D bounding boxes over a future time interval to form spatiotemporal motion envelopes. An overlap is determined based on whether these aggregated regions intersect. Depending on the time window in which overlap is detected, we assign three levels of severity: "HIGH" risk of overlap ($\Delta t = 1.5$\,s), "LOW" risk of overlap ($\Delta t = 3.0$\,s but not $1.5$\,s), or "NO" risk.

The formal definition is as follows. Let $\mathcal{B}_{\text{ego}}(t)$ and $\mathcal{B}_{\text{obj}}(t)$ denote the 2D bounding boxes of the ego vehicle and another object at time $t$. Define the aggregated motion regions:

\[
    \mathcal{A}_{\text{ego}}^{\Delta t}= \bigcup_{t \in [t_{\text{now}}, t_{\text{now}} + \Delta t]}\mathcal{B}_{\text{ego}}(t), \quad \mathcal{A}_{\text{obj}}^{\Delta t}= \bigcup_{t \in [t_{\text{now}}, t_{\text{now}} + \Delta t]}\mathcal{B}_{\text{obj}}(t)
\]

An overlap indicator function is defined as:

\[
    \mathcal{O}(\Delta t) =
    \begin{cases}
        1, & \text{if }\mathcal{A}_{\text{ego}}^{\Delta t}\cap \mathcal{A}_{\text{obj}}^{\Delta t}\neq \emptyset \\
        0, & \text{otherwise}
    \end{cases}
\]

Then the \textit{TrajOverlapTag} is assigned as:

\[
    \text{Tag}=
    \begin{cases}
        \text{HIGH}, & \text{if }\mathcal{O}(1.5\,\text{s}) = 1                                          \\
        \text{LOW},  & \text{if }\mathcal{O}(3\,\text{s}) = 1 \text{ and }\mathcal{O}(1.5\,\text{s}) = 0 \\
        \text{NO},             & \text{otherwise}
    \end{cases}
\]

Compared to the \textit{CollisionTag}, the proposed \textit{TrajOverlapTag} better accounts for protective braking behaviors exhibited by the driver, even in the absence of an actual collision if the ego vehicle were to maintain its velocity. Fig.~\ref{fig:exp_overlap} shows an example where \textit{TrajOverlapTag} correctly detects the guest vehicle triggering the ego vehicle's braking, while \textit{CollisionTag} does not.

The output of rule-based scenario tagging is summarized in table~\ref{tab:tag_results}, where each scenario can be represented as several matrix with $N$ agents and $T$ timestamps.

\begin{table}[htbp]
\centering
\caption{Output structure after rule-based scenario tagging.}
\label{tab:tag_results}
\begin{tabular}{lll}
\toprule
\textbf{Tag Name} & \textbf{Shape} & \textbf{Description} \\
\midrule
\text{TrackID}       & $[N,\ T]$     & \\
\text{ObjectCategory}       & $[N,\ T]$     & \\
\text{LongitudeTag}     & $[N{+}1,\ T]$ & includes Ego vehicle ($+1$). \\
\text{LatitudeTag}      & $[N{+}1,\ T]$ & includes Ego vehicle. \\
\text{HeadingTag}       & $[N,\ T]$     & \\
\text{PositionTag}      & $[N,\ T]$     & \\
\text{CollisionTag}     & $[N,\ T]$     & \\
\text{DistanceTag}      & $[N,\ T]$     & \\
\text{TrajOverlapTag}   & $[N,\ T]$     & \\
\bottomrule
\end{tabular}
\end{table}

\subsection{Key Object Identification (KeyIdent)}

In complex traffic scenarios, a large number of surrounding objects are typically detected within a single sequence, making it computationally exhaustive to perform \gls{llm}-based reasoning on every activity pair between the ego vehicle and each surrounding guest object. Furthermore, most surrounding objects are irrelevant to the ego vehicle’s behavior. To address this, we introduce a \gls{koi} module to filter only those candidates most likely to influence the ego vehicle’s decisions.

We specifically focus on braking behavior as an interaction indicator. When the longitude status of the ego vehicle is either "DECELERATING" or "STANDING\_STILL", and a surrounding object simultaneously exhibits either "LOW" or "HIGH" collision risk or "LOW" or "HIGH" trajectory overlap risk with the ego vehicle, the object is regarded as a key candidate responsible for the deceleration. This rule effectively filters out irrelevant objects and focuses \gls{llm} processing on those entities most likely to affect the ego vehicle’s behavior.

\subsection{LLM-based Scenario Understanding and Classification}

\begin{figure*}[ht]
    \centering
    \includegraphics[width=\linewidth]{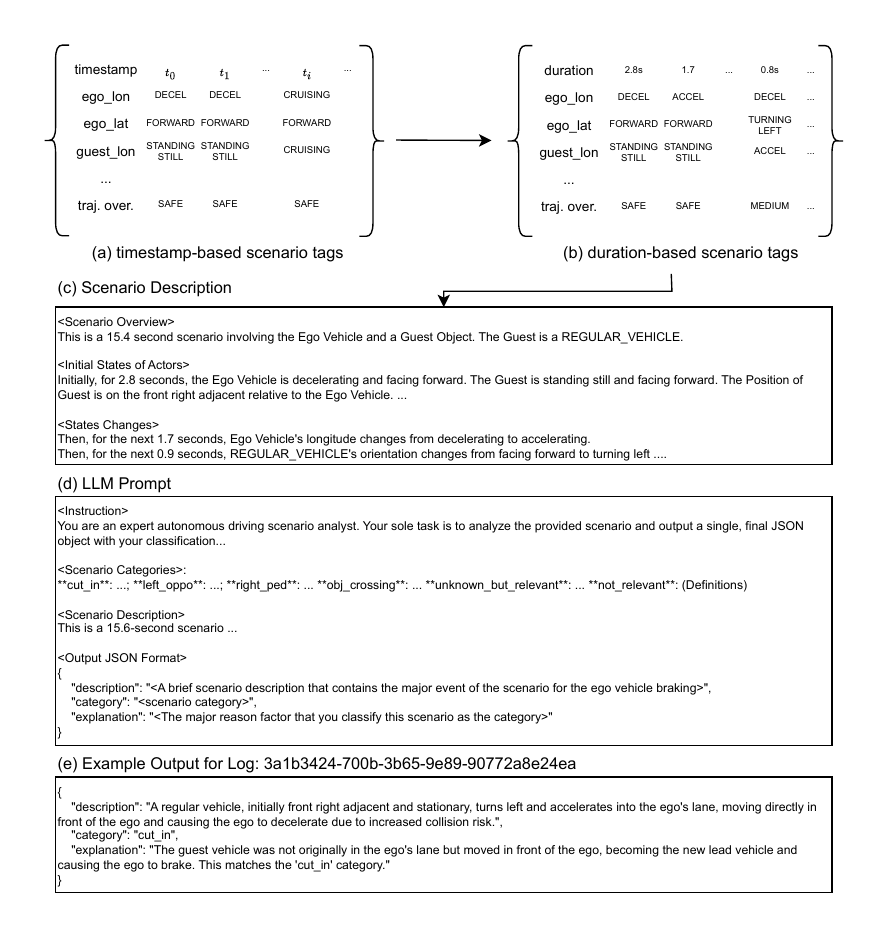}
    \caption{Building scenario description and prompt to LLM and an output example}
    \label{fig:sc_desc}
\end{figure*}

\subsubsection{Scenario Description Composition}

In the scenario understanding module, a \gls{llm} is employed to interpret and categorize scenarios based on predefined scenario categories. As illustrated in Fig.~\ref{fig:sc_desc}a~\ref{fig:sc_desc}b, the process begins by transforming timestamp-based scenario tag sequences into a duration-based representation. Consecutive timestamps with identical activity tags are merged into a single segment, and a duration is assigned to each resulting segment. 

Next, the duration-based segments are translated into a structured text description using a predefined template as shown in Fig.~\ref{fig:sc_desc}c. The generated description begins with an overview, including the total duration and the two primary actors involved in the scenario. This is followed by a detailed initialization of all activity tags. For subsequent segments, only the tags that change from the previous segment are included, along with their corresponding durations.

\subsubsection{LLM Prompt for Scenario Understanding}

The translated scenario description is combined with a predefined prompt and provided as input to the LLM, as shown in Fig.~\ref{fig:sc_desc}d. Following common prompt engineering principles~\cite{zhao2023survey}, the prompt consists of several components: a role definition and instruction, a list of known scenario categories, and an instruction to return the result in a fixed JSON format using specified keys.

Fig.~\ref{fig:sc_desc}e shows an output example, which includes the following fields:

\begin{table*}[htbp]
\centering
\caption{Statistics of annotated scenario categories in Argoverse 2 Sensor Dataset train split (700 data logs).}
\label{tab:annotation}
\begin{tabular}{l p{9cm} r}
\toprule
\textbf{Scenario Tag} & \textbf{Description} & \textbf{Count} \\
\midrule
cut\_in & An object cuts into the path of the ego vehicle. & 95 \\
left\_oppo & Ego vehicle turns left while an oncoming object approaches. & 5 \\
right\_ped & Ego vehicle turns right while a pedestrian is crossing. & 27 \\
obj\_cross & An object crosses in front of the ego vehicle. & 64 \\
ped\_cross & A pedestrian crosses in front of the ego vehicle. & 40 \\
lead\_brake & Sudden braking of the lead vehicle. & 56 \\
approach\_stop & Ego vehicle approaches a stationary or slow-moving object. & 91 \\
not\_relevant & Scenarios not relevant to ego vehicle behavior. & 313 \\
unknown\_but\_relevant & Hard-to-describe but relevant scenarios. & 36 \\
\midrule
\textbf{Total} & & \textbf{727} \\
\bottomrule
\end{tabular}
\end{table*}

\begin{itemize}
    \item \textbf{Scenario Description}: A rephrased version of the rule-based scenario description that highlights the key event(s) in the scenario.


    \item \textbf{Scenario Category}: A selected category from the provided list of known scenario types that best fits the current scenario. When none of them fit, output either \textbf{not\_relevant} or \textbf{unknown\_but\_relevant}.

    \item \textbf{Explanation}: The reasoning behind the chosen category, explaining why the scenario was classified as such.
\end{itemize}

Finally, the rephrased scenario description is encoded using an embedding model to obtain a vector representation for similarity-based retrieval.

These four \gls{llm}-generated enrichments are stored alongside the rule-based tags in the scenario database for downstream retrieval and analysis.


\subsection{Dual-Path Scenario Retrieval}

Since only a closed set of scenario categories is explicitly provided in the prompt, \gls{ood} scenarios are not directly assigned by the \gls{llm}. To address this limitation, we propose a dual-path scenario retrieval module that supports both structured and open-ended queries.

For queries corresponding to known categories, the system performs direct retrieval by matching the \texttt{Scenario Category} field labeled by the \gls{llm} in the scenario database. In addition, users can retrieve scenarios using free-form scenario description queries. To enable this, we implement an embedding model to generate a vector representation for each scenario description. Cosine similarity is applied to compute similarity scores between the query and stored scenarios.

This dual-path retrieval framework allows users to access both standard scenario categories and customized scenarios described in natural language, enhancing the system’s flexibility in handling both \gls{id} and \gls{ood} scenario retrieval.

\section{experimental Setup}

This section outlines the experimental setup, beginning with dataset preparation by annotating scenario categories on Argoverse 2, followed by implementation details of the rule-based and \gls{llm}-based module. 

\subsection{Dataset}


Our method is evaluated on the Argoverse 2 Sensor dataset~\cite{wilson2023argoverse}, which consists of 1000 diverse urban driving logs featuring high-definition maps, time-synchronized multimodal sensor data (LiDAR, cameras, radar), and 3D object track annotations. We curate a set of high-level scenario category annotations focusing on challenging driving behaviors on the training data split (700 driving logs). As shown in table~\ref{tab:annotation}, the scenarios are classified into nine categories, including seven specific scenario types as well as \textit{not\_relevant} and \textit{unknown\_but\_relevant}. Each annotation consists of three elements: \textit{log\_id}, \textit{guest\_id}, and \textit{scenario\_category}. It should be noted that one data log may contain multiple scenarios, each involving different guest objects.

\subsection{Implementation Details}

We follow the rule-based scenario tagging setup from~\cite{guo2023scenario}, with several refinements. The \textit{LongitudeTag} remains unchanged. For \textit{LatitudeTag}, we subdivide "TURNING" into "VEERING" and "TURNING": an accumulated turning angle greater than $0.17$ radians is labeled as "TURNING," while an angle between $0.05$ and $0.17$ radians is labeled as "VEERING." For \textit{CollisionTag}, \gls{ttc} greater than 3 seconds is labeled "NO" risk, between 1.5 and 3 seconds as "LOW" risk, and less than 1.5 seconds as "HIGH" risk. For \textit{PositionTag} and \textit{HeadingTag}, angular thresholds are equally divided based on the number of defined classes. For \textit{DistanceTag}, overlapping expanded bounding boxes are classified as "VERY\_CLOSE"; distances below 10 meters as "CLOSE"; distances from 10 to 50 meters as "MEDIUM"; and distances over 50 meters as "FAR." Implementation details on \textit{TrajOverlapTag} are provided in Section~\ref{sec:trajoverlap}.

For \gls{llm}-based modules, we evaluated three \gls{llm} models: GPT-4.1~\cite{achiam2023gpt}, Gemini-2.5-Flash~\cite{team2023gemini} and Qwen3-32B~\cite{yang2025qwen3}. We primarily report the results of the best-performing model (gemini-2.5-flash-preview-05-20), along with a comparative analysis across all three models. For text embedding, we employ the opensource nomic embed model~\cite{nussbaum2024nomic}. 

To ensure consistent and deterministic outputs from the \gls{llm}, we set the sampling parameters as follows: temperature = 0, top\_p = 1.0, and top\_k = 1. This configuration disables randomness in token sampling and makes the model generate the most likely output sequence given the prompt.

\section{Evaluation}

We evaluate the effectiveness of our proposed method in comparison to the baseline rule-based tagging and retrieval approach introduced by~\cite{guo2023scenario}. 

Since our proposed framework is in two stages: \gls{koi} followed by \gls{llm}-based scenario understanding, we organize the evaluation into four parts: (1) evaluation of the \gls{koi}; (2) evaluation of scenario retrieval for known \gls{id} scenario categories; (3) evaluation of unknown scenario retrieval by query description; and (4) comparison of performance with different \gls{llm} models.

\subsection{Evaluation for Key Object Identification (KeyIdent)}


\begin{table}[htbp]
\centering
\caption{Evaluation for \gls{koi} module}
\label{tab:eval_keyident}
\begin{tabular}{l c c}
\toprule
\textbf{Category} & \textbf{Precision} & \textbf{Recall} \\
\midrule
cut\_in        & - & 0.96 \\
left\_oppo      & - & 0.80 \\
right\_ped  & - & 0.74 \\
obj\_crossing   & - & 0.86 \\
\hline
Overall & 0.37 & 0.86 \\
\bottomrule
\end{tabular}
\end{table}
\begin{table}[htbp]
\centering
\caption{Evaluation for four known \gls{id} scenario categories.}
\label{tab:eval1}
\begin{tabular}{lccc}
\toprule
\textbf{Scenario Category} & \textbf{Precision} & \textbf{Recall} & \textbf{F1} \\
 & \multicolumn{3}{c}{\textbf{Baseline~\cite{guo2023scenario} / Ours}} \\
\midrule
cut\_in & 0.18 / \textbf{0.48} & \textbf{0.97} / 0.85 & 0.31 / \textbf{0.61} \\
left\_oppo & \textbf{0.14} / 0.12 & 0.25 / \textbf{1} & 0.18 / \textbf{0.22} \\
right\_ped & \textbf{0.60} / 0.58 & 0.375 / \textbf{0.63} & 0.46 / \textbf{0.60} \\
obj\_crossing & \textbf{0.68} / 0.31 & 0.29 / \textbf{0.69} & 0.41 / \textbf{0.43} \\
\midrule
Overall & 0.22 / \textbf{0.39} & 0.66 / \textbf{0.78} & 0.33 / \textbf{0.52} \\\bottomrule
\end{tabular}
\end{table}
As shown in Table~\ref{tab:eval_keyident}, the high recall of 0.86 ensures that most relevant pairs are captured. Although the precision is relatively low at 0.37, it is acceptable within our framework. The second-stage \gls{llm} module is specifically designed for scenario understanding and classification, allowing it to filter out false positives. Since the \gls{koi} module does not differentiate between scenario categories, category-wise precision scores are not reported.

\subsection{Evaluation for \gls{id} Scenarios}

We first evaluate the scenario classification performance on four categories that are explicitly given in the prompt, as shown in Fig.~\ref{fig:sc_desc}d. To ensure a fair comparison, we carefully adjust the baseline retrieval filter condition to maximize its F1 score, making a trade-off between precision and recall.


As shown in Table~\ref{tab:eval1}, our \gls{llm}-based method consistently outperforms the baseline in terms of F1 score across all categories as well as overall. While precision is lower in some categories, such as \textit{obj\_crossing}, it shows much stronger recall to ensure comprehensive scenario coverage. We notice the low precision on \textit{left\_oppo} and \textit{obj\_crossing} and analyze them in Section~\ref{sec:fc_cases}.




\subsection{Evaluation for \gls{ood} Scenario Description Retrieval}

To evaluate the second retrieval path, which retrieval scenarios by a given \gls{ood} scenario description query, we evaluate three additional scenario categories that are not provided in the prompt: \textit{approach\_stop}, \textit{lead\_brake}, and \textit{ped\_crossing}. We report precision at top-10 (P@10) and recall at top-50 (R@50) based on similarity scores in Table~\ref{tab:eval3}.

The results show that similarity-based scenario retrieval generally performs lower than \gls{id}-based retrieval. However, given that it operates in a zero-shot setting, the performance remains acceptable. In particular, the recall scores for \textit{lead\_brake} and \textit{ped\_crossing} reach 0.5 and 0.67 at Top-50, respectively. The performance on \textit{approach\_stop} scenarios is slightly lower, likely because these scenarios always involve non-emergent braking, which may lead the \gls{llm} to underestimate their relevance during retrieval.



\begin{table}[htbp]
\centering
\caption{Evaluation for Similarity-based Scenario Retrieval}
\label{tab:eval3}
\begin{tabular}{lp{3cm}ccc}
\toprule
\textbf{\makecell{Scenario \\ Category}} & \textbf{\makecell{
Query\\Description}} & \textbf{P@10} & \textbf{R@50} \\
\midrule
approach\_stop & "The ego vehicle approaches a stopped or slow-moving vehicle ahead, requiring careful deceleration to maintain a safe distance." & 0.30 & 0.31  \\
\midrule
lead\_brake & "The lead vehicle suddenly brakes, requiring the ego vehicle to brake in order to avoid a collision." & 0.40 & 0.50  \\
\midrule
ped\_crossing & "A pedestrian crosses in front of the ego vehicle, requiring the ego vehicle to brake or yield to avoid a collision." & 0.30 & 0.67 \\
\bottomrule
\end{tabular}
\end{table}
\begin{table}[h!]
\centering
\caption{Comparison of method with different LLM models}
\label{tab:eval_llm}
\begin{tabular}{lccc}
\toprule
\textbf{Scenario Category} & \textbf{Precision} & \textbf{Recall} & \textbf{F1} \\
\midrule
Qwen3-32B  & 0.29 & 0.60 & 0.39\\
GPT-4.1 & 0.37 & 0.63 & 0.46 \\
gemini-2.5-flash-preview-05-20  & \textbf{0.39} & \textbf{0.78} & \textbf{0.52}\\
\bottomrule
\end{tabular}
\end{table}

\subsection{Comparison of LLM Models}

We report the performance comparison of three \gls{llm} models: GPT-4.1, Gemini-2.5-Flash, and Qwen3-32B. The latter two are reasoning models, which require longer inference time. Among the three, Gemini-2.5-Flash achieves the best overall performance, with an F1 score of 0.52 and a recall of 0.78. In contrast, Qwen3-32B shows the lowest performance, primarily due to its limited token window, which is often exceeded by long scenario descriptions. All three models outperform the baseline method. This confirms the effectiveness of the proposed method.

\subsection{Failure Case Analysis}
\label{sec:fc_cases}
\begin{figure*}[htbp]
    \centering
    \begin{subfigure}[t]{0.30\textwidth}
        \centering
        \tcbset{colframe=black,boxrule=0.5pt,arc=0mm,boxsep=0mm,left=0mm,right=0mm,top=0mm,bottom=0mm}
        \begin{tcolorbox}
            \includegraphics[width=\textwidth]{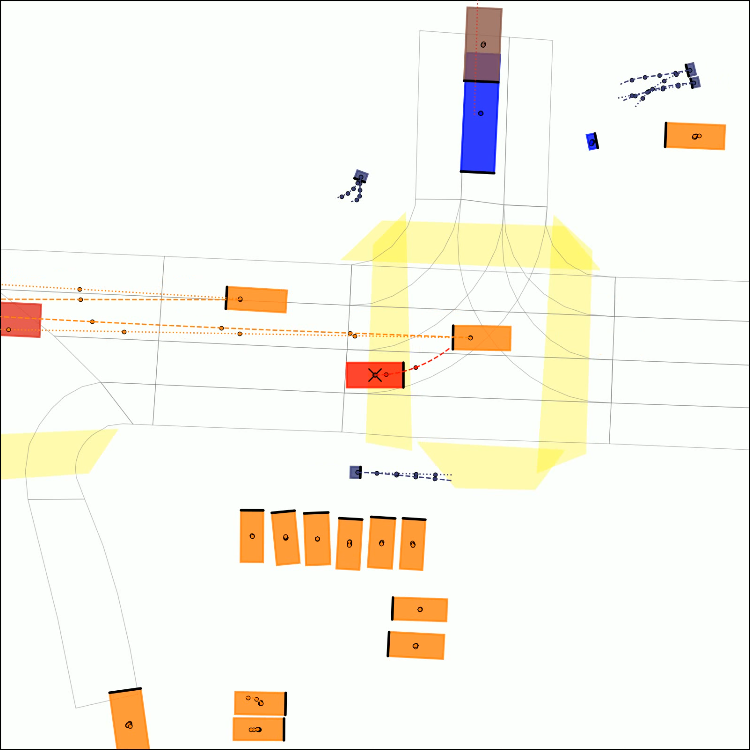}
        \end{tcolorbox}
        \caption{\textit{left\_oppo}: The ego vehicle starts braking before it turns left due to the oncoming vehicle. This scenario is mistakenly classified as \textit{obj\_crossing}."}
        \label{fig:fc_left_oppo}
    \end{subfigure}
    \hfill
    \begin{subfigure}[t]{0.30\textwidth}
        \centering
        \tcbset{colframe=black,boxrule=0.5pt,arc=0mm,boxsep=0mm,left=0mm,right=0mm,top=0mm,bottom=0mm}
        \begin{tcolorbox}
            \includegraphics[width=\textwidth]{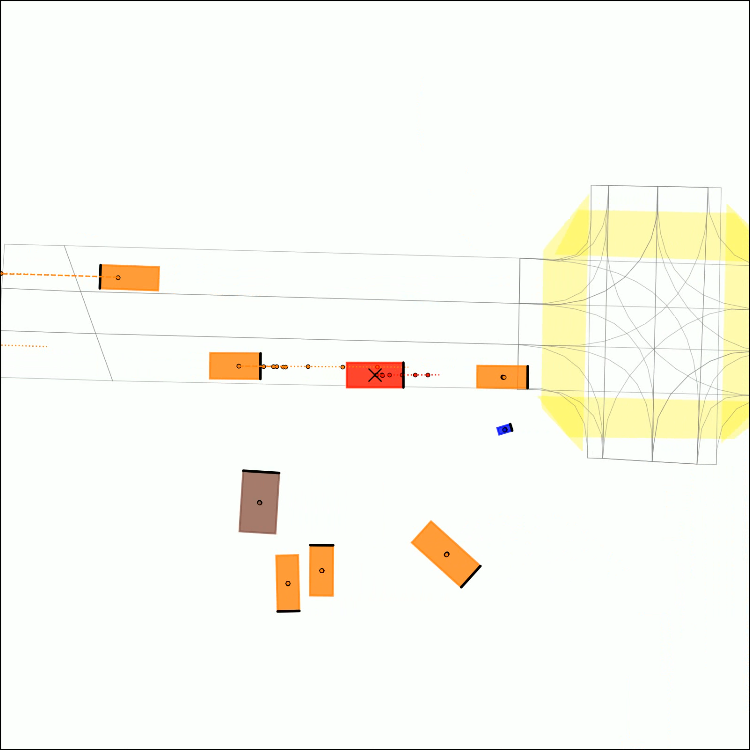}
        \end{tcolorbox}
        \caption{\textit{approach\_stop}: The ego vehicle approaches a stationary vehicle and initiates early braking. No collision risk detected.}
        \label{fig:fc_approach_stop}
    \end{subfigure}
    \hfill
    \begin{subfigure}[t]{0.30\textwidth}
        \centering
        \tcbset{colframe=black,boxrule=0.5pt,arc=0mm,boxsep=0mm,left=0mm,right=0mm,top=0mm,bottom=0mm}
        \begin{tcolorbox}
            \includegraphics[width=\textwidth]{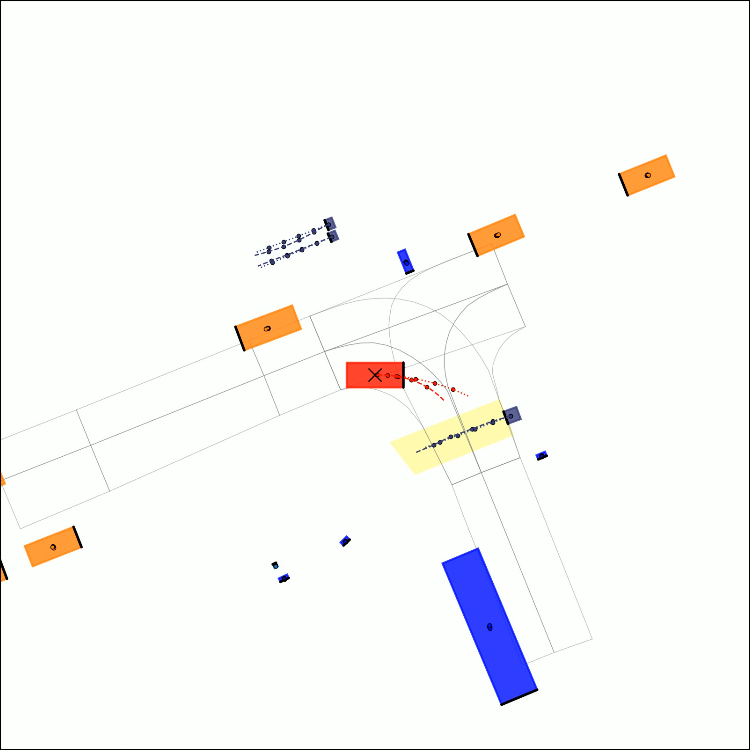}
        \end{tcolorbox}
        \caption{\textit{right\_ped}: The ego vehicle brakes early and stops in front of the crosswalk. No collision risk detected.}
        \label{fig:fc_right_ped}
    \end{subfigure}
    \caption{Visualization of failure cases in scenario extraction. Red: ego vehicle; orange: guest vehicle; grey: pedestrian; dashed line: actual future trajectory; dotted line: predicted future trajectory.}
    \label{fig:failure_cases}
\end{figure*}

Despite the promising results, we observe that certain braking events remain challenging to interpret correctly. In particular, the scenario category \textit{left\_oppo} shows the lowest performance among the four scenario types. Fig.~\ref{fig:fc_left_oppo} illustrates one example: while the ego vehicle is stopped in a waiting area for a left turn, an oncoming vehicle approaches and crosses in front of the ego vehicle. The absence of map information makes it difficult for the \gls{llm} to recognize intersections, causing confusion between \textit{left\_oppo} and \textit{obj\_crossing} scenarios, as both involve a vehicle crossing in front of the ego vehicle with similar patterns.

Another common failure case occurs during the \gls{koi}, particularly in the scenarios \textit{right\_ped} and \textit{approach\_stop}. As illustrated in Fig.~\ref{fig:fc_approach_stop} and Fig.~\ref{fig:fc_right_ped}, the ego vehicle applies only gentle, anticipatory braking rather than a hard brake, as the potential collision is foreseen early. Consequently, neither collision risk nor trajectory overlap is detected. This suggests that the proposed method is more effective in handling critical braking scenarios, while it remains limited in capturing softer or anticipatory braking cases.

\section{Conclusion}

In this work, we present a novel \gls{llm}-based scenario extraction and reasoning framework for identifying the causes of braking events in driving. By translating low-level BUS data into structured natural language descriptions, our method enables \gls{llm}-powered scenario understanding and classification. We introduce a dual-path retrieval method that supports both \gls{id} and \gls{ood} scenario retrieval. Furthermore, we provide scenario annotations on the Argoverse 2 sensor dataset to facilitate evaluation. Experimental results show that our approach not only outperforms baseline rule-based methods but also demonstrates strong zero-shot generalization to previously unseen \gls{ood} scenarios. 

{
    \bibliographystyle{IEEEtran}
    \small
    \bibliography{references}
}

\end{document}